\documentclass[letterpaper, 11 pt, conference]{ieeeconf}  
\usepackage{graphicx} 
\usepackage{multirow}
\usepackage{booktabs}
\usepackage[font=small]{caption}
\usepackage{subcaption}

\IEEEoverridecommandlockouts                              
\usepackage[utf8]{inputenc}
\usepackage[T1]{fontenc}

\title{\LARGE \bf Predicting Students' Exam Scores \\ Using Physiological Signals}

\author{ Willie Kang$^{*1, 2}$, Sean Kim$^{*1, 3}$, Eliot Yoo$^{*1, 4}$, and Samuel Kim$^{1}$ 
\\ \\
$^1$ IF Research Lab, La Palma, CA, USA \\
$^2$ El Toro High School, Lake Forest, CA, USA \\
$^3$ Oxford Academy, Cypress, CA, USA \\
$^4$ Cypress High School, Cypress, CA, USA 
\\
{\tt\small \{wildmanwillie25, seankim.hahjean, philliot1304\}@gmail.com} \\{\tt\small sam@ifresearchlab.com}
\thanks{$^{*}$ These authors contributed equally. The names are listed in an alphabetical order.}
}

\begin{document}

\maketitle
\begin{abstract}
While acute stress has been shown to have both positive and negative effects on performance, not much is known about the impacts of stress on students’ grades during examinations. To answer this question, we examined whether a correlation could be found between physiological stress signals and exam performance. We conducted this study using multiple physiological signals of ten undergraduate students over three different exams. The study focused on three signals, i.e., skin temperature, heart rate, and electrodermal activity. We extracted statistics as features and fed them into a variety of binary classifiers to predict relatively higher or lower grades. Experimental results showed up to 0.81 ROC-AUC with $k$-nearest neighbor algorithm among various machine learning algorithms. 
\end{abstract}

\section{INTRODUCTION}
College students are prone to stress due to the highly transitional and demanding nature of their lives, which may be because of rigorous academic requirements, an unfamiliar environment, and separation from home. Academic stress is a regular part of the lives of students, and may result from pressures to perform, perceptions of workloads and exams, and time restraints \cite{bedewy_gabriel_2015}. Failure to cope with such high stress can lead to various negative effects. Severe academic stress decreases academic performance and hinders the ability to study effectively \cite{Khan2018, Sohail2013}. Overall, stress has been shown to negatively impact sleep quality, well being, and affectivity, which in turn negatively impacts general health \cite{wunsch_kasten_fuchs_2017}. 

Additionally, students may experience more severe issues during examination season. This period is often marked by high stress and anticipation, with numerous important projects, papers, and exams all colliding. During this time, sleep quality has been shown to decrease and caffeine consumption has been shown to increase \cite{zunhammer_eichhammer_busch_2014, campbell_soenens_beyers_vansteenkiste_2018}. 

Students are also adversely impacted by test anxiety. Higher levels of cognitive test anxiety have been associated with significantly lower test scores \cite{cassady_johnson_2002}. A study of nursing students has also shown that test anxiety causes physical, emotional, and cognitive detriments, which hinders academic success \cite{shapiro_2014}. There also exists an inverse relationship between test anxiety and grade point average in both graduate and undergraduate students \cite{chapell_blanding_silverstein_takahashi_newman_gubi_mccann_2005}. 

Exam stress and anxiety is a significant problem that affects all students. Working on this issue can lead to not only academic improvements, but physical and mental health benefits. Being able to predict exam performance through common physiological signals that correlate with stress can serve as a useful tool to help address the issue of test anxiety. Therefore, this study aims to look at the viability of predicting exam scores with physiological signals using machine learning algorithms.

\begin{figure*}[t!]
\centering
  \begin{subfigure}[]{0.3\linewidth}
    \includegraphics[width=\linewidth]{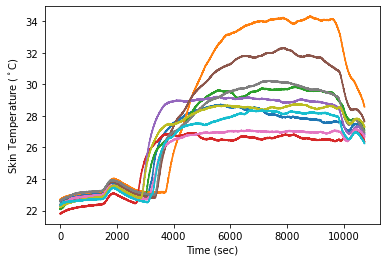}
     \caption{Skin temperature - Midterm 1}     
  \end{subfigure}
  \begin{subfigure}[]{0.3\linewidth}
    \includegraphics[width=\linewidth]{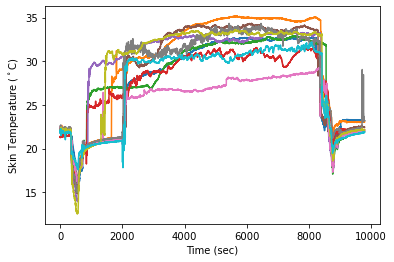}
     \caption{Skin temperature - Midterm 2}
  \end{subfigure}
  \begin{subfigure}[]{0.3\linewidth}
    \includegraphics[width=\linewidth]{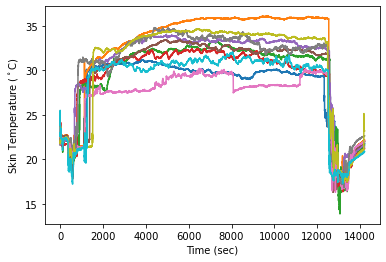}
    \caption{Skin temperature - Final}
  \end{subfigure}
  \begin{subfigure}[]{0.3\linewidth}
    \includegraphics[width=\linewidth]{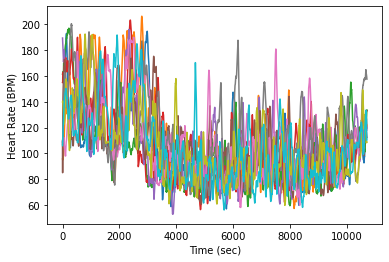}
     \caption{Heart Rate - Midterm 1}     
  \end{subfigure}
  \begin{subfigure}[]{0.3\linewidth}
    \includegraphics[width=\linewidth]{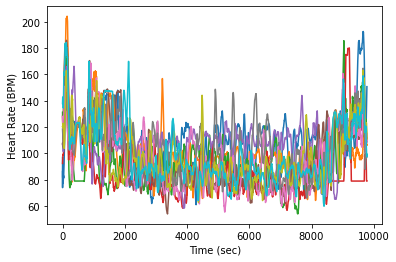}
     \caption{Heart Rate - Midterm 2}
  \end{subfigure}
  \begin{subfigure}[]{0.3\linewidth}
    \includegraphics[width=\linewidth]{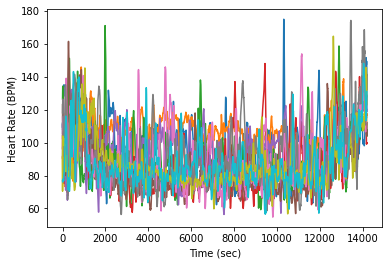}
    \caption{Heart Rate - Final}
  \end{subfigure}
  \begin{subfigure}[]{0.3\linewidth}
    \includegraphics[width=\linewidth]{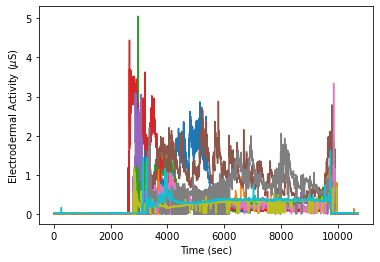}
     \caption{Electrodermal activity - Midterm 1}     
  \end{subfigure}
  \begin{subfigure}[]{0.3\linewidth}
    \includegraphics[width=\linewidth]{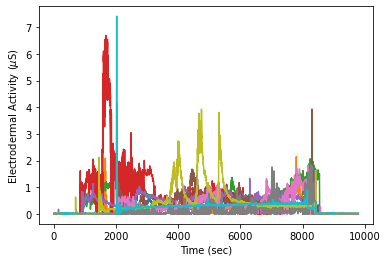}
     \caption{Electrodermal activity - Midterm 2}
  \end{subfigure}
  \begin{subfigure}[]{0.3\linewidth}
    \includegraphics[width=\linewidth]{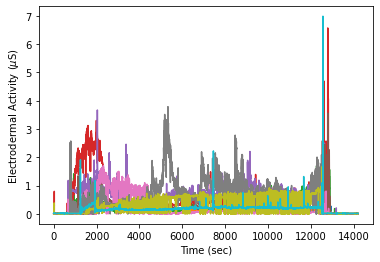}
    \caption{Electrodermal activity - Final}
  \end{subfigure}
  \caption{Physiological signals of the individual students during exams. }
  \label{fig:signals}
\end{figure*}

\section{Procedure}
\subsection{Data Source}
The data we used was collected from a study conducted at the University of Houston on eleven undergraduate students (nine males, two females) who were tracked across three major exams: two midterms and a final exam~\cite{data2}. The students wore E4 wristbands that measured skin conductance, electrodermal activity (EDA), heart rate, blood volume pulse, skin surface temperature, inter-beat interval, and accelerometer data. 

Of the eleven participants, one student was provided additional accommodations due to the University of Houston disability accommodation guidelines. Data from this participant was discarded as it involved a factor not consistent with the other participants. See~\cite{data1} for more details. 

For our research, we chose to incorporate skin temperature, heart rate, and EDA measurements. Figure~\ref{fig:signals} shows the selected physiological signals of individual students collected during different examinations.

\subsection{Pre-Processing}

Firstly, we synchronized all the measurements aligned at the same timestamp. Since the data was collected in an asynchronized manner, we dropped out any measurements that are outside of common time periods. 

Secondly, we found some outliers and missing values in measurements. Therefore, we applied a filtering method, moving average low-pass filter to be specific, to remove possible noise and outliers. 

Lastly, the physiological signals can be influenced by personal biases and environmental factors. For example, individual skin temperatures can be influenced by the room temperature and some students can have innately higher heart rates then the others. To mitigate these biases, we normalized the data before inputting the data into the machine learning algorithms. The normalization was done both on a student-level and a test-level.  
We used the z-normalization so that individual instruments have zero means and unit standard deviations, i.e.
\begin{equation}
x(t) = \frac{x(t) - \mu}{\sigma}
\end{equation}
where $x(t)$ represents measured value of the instrument at time $t$, and $\mu$ and $\sigma$ are the notation for the average and the standard deviation of the measurement over time, respectively.

\subsection{Feature Extraction}
As described earlier, we used skin temperature, heart rate, and EDA of the students. 
After the pre-processing, we extracted the statistics of a physiological signal as feature vectors to the instrument during an exam. The statistics consist of mean, standard deviation, minimum, maximum, and median (the feature dimension is 5). Then, we concatenate all the features to create one super-vector to represent overall physiological behaviors during the exam (the dimension of the super-vector is 15). 

Since one student takes three different exams, i.e. two midterms and one final, each student will have three different physiological behavior features and corresponding test scores.

\begin{figure}[t!]
  \centering
      \includegraphics[width=8cm]{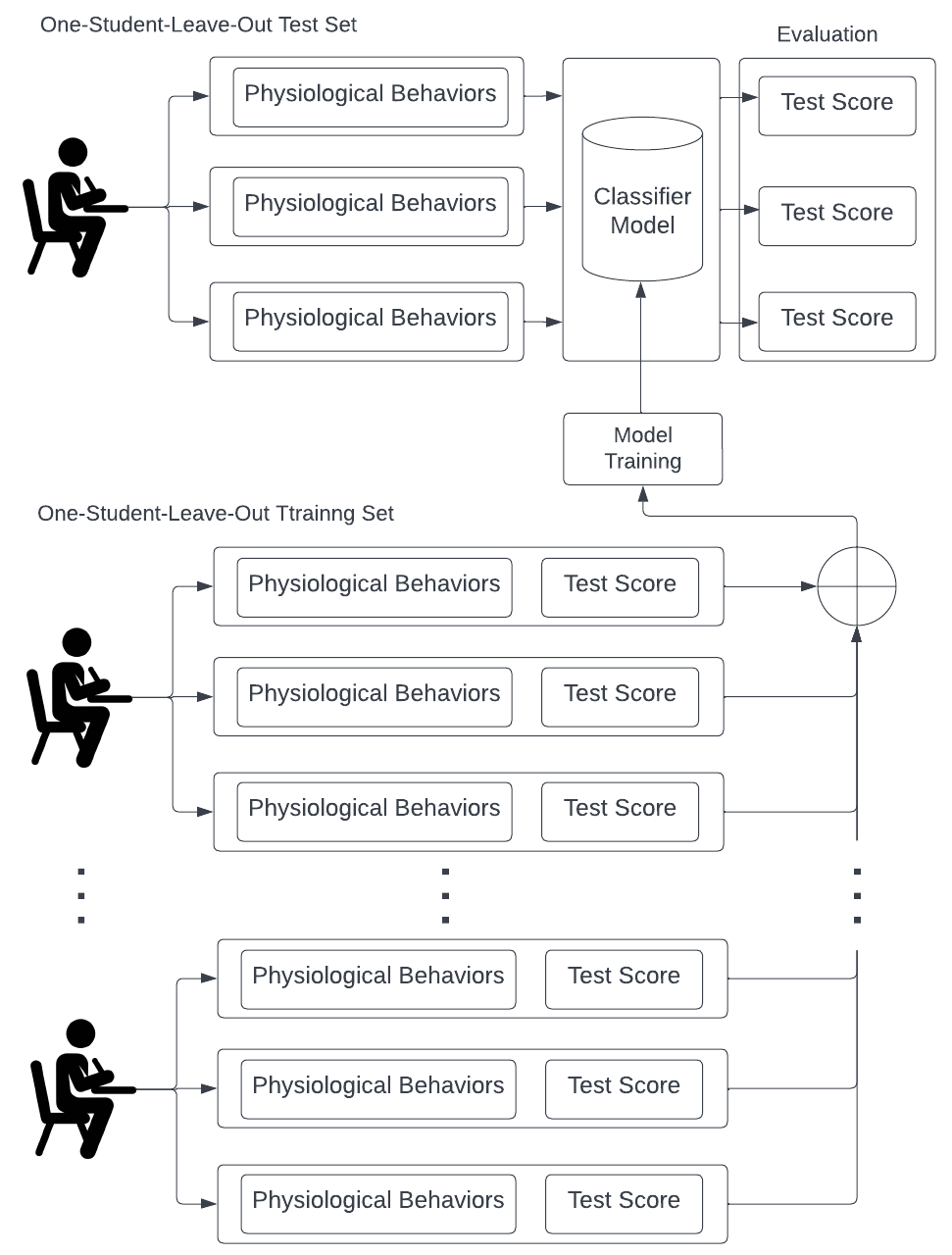}
  \caption{Basic diagram of each validation step for one-student-leave-out experimental setup.}
  \label{fig:diagram}
\end{figure}
\section{Experiments}
\subsection{Experimental Setup}
We used all the features regardless of exam types so that each student has three different scores and corresponding physiological features. The train and test sets were split in a one-student-leave-out way which means nine students would be used to train the classifier and the other student would be used to test it. This creates 10-fold cross-validation, and each validation task consists of 27 training samples and 3 test samples. Figure~\ref{fig:diagram} illustrates this scenario as a simple diagram.

We designed the experiments as binary classification tasks. In this regard, we built models to classify whether students received a score higher than 80

We repeated the experiments 10 times so that we can get the average performance of individual machine learning algorithms.

\subsection{Classifiers}
Multiple machine learning models were used. Using a diverse amount of classifiers allows for the various algorithms to search for a correlation between the stress signal values and the performance of the student. These machine learning models were the Random Forest (RF, with a gridsearch technique for best parameters in each validation task), Stochastic Gradient Descent (SGD, with log-loss), Support Vector Machine (SVM, with RBF kernel and $C=1$), and $k$-nearest neighbor (KNN, with $k=5$) classifiers.

\subsection{Results}
\begin{figure}[t!]
  \centering
      \includegraphics[width=8cm]{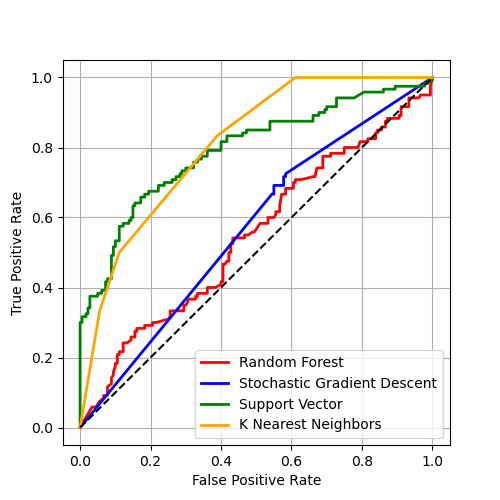}
  \caption{ROC curves with various machine learning algorithms.}
  \label{fig:auc score}
\end{figure}

Figure~\ref{fig:auc score} and Table~\ref{table:auc score} show the results of the binary classification tasks in terms of ROC-AUC using various machine learning algorithms. 
Overall, the KNN gave the best results with a 0.81 ROC-AUC on average in the relationship between stress levels and high scoring on exams. This classifier shows that there exists a correlation between stress and test scores that could be further investigated to find a stronger relation on how stress levels can affect the performance of a student. The SVM Classifier produced the second best results with a 0.80 ROC-AUC in the relationship between stress and exam scores. This further shows that there is a considerable correlation between stress and scores. 

On the other hand, RF and SGD did not yield sufficient ROC-AUC scores, which indicate that those machine learning algorithms are not performing well enough to model the relationship between physiological behaviors and test scores.

\subsection{Limitations}
One limitation of our study is the small number of statistics extracted from the chosen physiological signals during feature extraction. We only utilized basic statistics as features. Using more comprehensive features may serve to better map the physiological signals to the exam scores. Furthermore, analyzing a larger dataset may help improve the accuracy of results.

\begin{table}[t!]
\centering
\begin{tabular}{c|c|c|c|c}
\hline 
 & RF & SGD & SVM & KNN \\
\hline \hline
\multirow{2}{*}{ROC-AUC} & 0.54 & 0.56 & 0.80 & 0.81 \\
         & (0.09) & (0.06) & (0.06) & (0.00) \\
\hline
\end{tabular}
 \caption{Average ROC-AUC scores (standard deviation) with various machine learning algorithms.}
\label{table:auc score}
\end{table}

\section{Conclusion}
The present research examined how stress affects academic performance through physiological signals. 
The results of this study support the initial hypothesis, suggesting a correlation between stress and exam results. These preliminary results have multiple implications for future research and further developments in the field. By looking at stress measurements, we can formulate strategies to maximize academic performance by looking at optimal levels of stress. Additionally, we can identify what factors have the greatest impact on stress and academic performance. Certain physiological signals may have detrimental effects on performance as they increase, while others may function in the inverse. Measurements that have the greatest impact on academic performance can be further investigated through various research and testing. The hypothesis attempted to form a general correlation based on the limited data and information available, but the opportunities for further improvement and different program creation are abundant. We expect future works to use this research as a foundation for more elaboration and growth within the fields of academics and stress. 

\bibliographystyle{IEEEbib}
{\bibliography{mybib}}




\end{document}